[a]biswasa@ornl.gov
[b]sergei2@utk.edu

# Combining Variational Autoencoders and Physical Bias for Improved Microscopy Data Analysis

Arpan Biswas[1,a], Maxim Ziatdinov[1,2] and Sergei V. Kalinin[3,b]

[1] Center for Nanophase Materials Sciences, Oak Ridge National Laboratory, Oak Ridge, TN 37831
[2] Computational Sciences and Engineering Division, Oak Ridge National Laboratory, Oak Ridge, TN 37831
[3] Materials Science and Engineering, University of Tennessee, Knoxville, TN 37996

Electron and scanning probe microscopy produce vast amounts of data in the form of images or hyperspectral data, such as EELS or 4D STEM, that contain information on a wide range of structural, physical, and chemical properties of materials. To extract valuable insights from these data, it is crucial to identify physically separate regions in the data, such as phases, ferroic variants, and boundaries between them. In order to derive an easily interpretable feature analysis, combining with well-defined boundaries in a principled and unsupervised manner, here we present a physics augmented machine learning method which combines the capability of Variational Autoencoders to disentangle factors of variability within the data and the physics driven loss function that seeks to minimize the total length of the discontinuities in images corresponding to latent representations. Our method is applied to various materials, including NiO-LSMO, $BiFeO_3$, and graphene. The results demonstrate the effectiveness of our approach in extracting meaningful information from large volumes of imaging data. The fully notebook containing implementation of the code and analysis workflow is available at https://github.com/arpanbiswas52/PaperNotebooks

## Introduction

[a]biswasa@ornl.gov
[b]sergei2@utk.edu

Over the last two decades, electron and scanning probe microscopies have become the instrument of choice for the visualization of the atomic and mesoscale structures and functionality of materials.[1–4] In electron microscopy, high resolution images of 2D materials,[5,6] catalysts,[7] functional oxides,[8] metals, and semiconductors have become routine.[9]Scanning probe microscopy have provided images of the surface topography, functional properties and forces,[10,11] ferroelectric[12] and magnetic domains.[13,14] High resolution imaging in liquids and ultra-high vacuum has provided spectacular images of self-assembled monolayers,[15] polymer chains,[16,17] and electronic density of states and order parameters.[18–20]

The broad availability of imaging data across length scales has further necessitated generating of physically relevant insight on materials microstructure. These analyses are closely tied to the nature of materials systems. For example, in high-resolution STEM images of oxide materials we seek to identify phases and ferroic variants and delineate the grain boundaries between them. In STM images of quantum materials we seek to identify individual defects and the characteristic wavelength of the symmetry breaking distortions, etc. Traditionally, these analyses were based on the uses of the techniques based on the Fourier transforms, recently combined with the linear decomposition methods such as principal component analysis. Alternatively, a broad set of analysis methods can be introduced once special points such as atomic locations are identified and the atomic neighborhoods[18–20] or image patches centered at these points are analyzed via linear decomposition methods.

The broad availability of deep learning methods over the last five years has significantly broadened the opportunities for image analysis.[21] The supervised machine learning methods have been applied for atomically resolved STEM[22–24] and STM[25–28] images. However, by definition supervised and semi-supervised methods rely on the human operator to provide the labeled examples. As such, they can discover only known structures and behaviors. The second limitation of the supervised methods is their vulnerability to the out of distribution drift effects.[29] Here, the network trained using examples acquired under one imaging conditions such as resolution, sampling, or other microscope-specific parameters will dramatically degrade performance under different conditions.

These considerations necessitate the development of the unsupervised machine learning methods for the image analysis. For systems with a well-defined crystallographic lattice, methods based on linear decompositions and variational autoencoders were shown to be highly efficient. The development of the rotationally invariant autoencoders extended these approaches to general orientation and discovery of chemical transformation pathways in disordered systems.[30–33] Also, such rotationally invariant autoencoders have been tuned with Bayesian optimization to maximize uncovering of the features from complex microscopic data.[34] However, for the time being these methods relied on the descriptors centered at the individual atomic units, whereas unsupervised discovery allowing for translation invariance has remained far more complex. Overall, all analysis to date relied on the local properties of the system, whereas the inferential biases associated with possible spatial structures, periodicities, etc. were limited.

Here, we introduce the approach for the analysis of the atomic and mesoscopic images that combines the capability of variational autoencoders to discover the factors of variability in the data under the invariances and the physics-based global criteria including existence and minimization of length of phase boundaries, presence of periodic orderings , etc. This is accomplished via the custom VAE approach combining classical reconstruction and Kullback-Leibler (KL) losses and physics-based reconstruction loss imposing physical constraints on the global characteristics of the latent images such as interface length or periodicities. This approach was implemented for the open data sets of ferroelectric domains in BiFeO3,[35] LSMO-NiO interface,[36] and graphene.[37]

[a]biswasa@ornl.gov
[b]sergei2@utk.edu

## 1. Methodology

Here, we discuss on the key components focused for this paper, namely, Variational Autoencoder (VAE) and Shift (or translational)-invariant variational autoencoders (sh-VAE). Finally, we discuss the formulation of the customized physics driven loss function, and its integration to sh-VAE, to build the physics-driven physics driven loss function, and its integration to sh-VAE, to build the physics-driven VAE model (phy-VAE) for better image analysis and discoveries. It is worthy to mention that in this paper, though we have focused the phy-VAE as the modification of sh-VAE, the same can be easily build (with similar proposed approach) on any other VAE models.

### *Variational Autoencoder (VAE) to Shift-Invariant Variational Auto-Encoder (sh-VAE)*

A variational autoencoder (VAE)[38] is a deep generative probabilistic model that belongs to the family of probabilistic graphical models and variational Bayesian methods. The two primary components of any VAE models are the encoder and decoder. Here, given an N-dimensional input, $X$, the encoder projects the input into a reduced n-dimensional (generally n = 2 or 3) latent variables, $Z$, following a distribution $p(Z|X)$. Then, with any sampling in the latent space, the decoder reconstructs the n-dimensional latent variables into the respective estimated N-dimensional input, $\overline{X}$, following a distribution $p(\overline{X}|Z)$. This encoding-decoding process of the VAE models need to be optimized such that the models can best learn the training data with minimizing the loss of information. The loss function of a VAE, $\mathcal{L}_{vae}$, which we minimize, can be defined as the sum of reconstruction error and Kullback–Leibler (KL) divergence and can be mathematically written as Eq. 1. Here, the reconstruction error can be chosen as mean square error, the cross-entropy error, etc. The KL divergence[39,40] $D_{KL}(p(z|x)||p(z))$ is the distance loss between the prior $p(z)$ distribution (usually chosen as standard Gaussian) and the posterior $p(z|x)$ distributions of the latent representation from data. VAE models have been implemented to materials systems on various tasks like classification, feature or pattern recognition, prediction etc though unsupervised, semi-, or supervised learning.[41–45]

This general or vanilla VAE is not robust to certain factors of variability in an image, such as rotation, translation, scale of the data, during the training process. However, the extension of this vanilla VAE has been attempted to develop VAE models which enforce rotational, translation or scale in-variances individually or together, while learning the individual continuous latent representations or jointly learning both continuous and discrete latent representation from the data, as implemented in pyroVED package in Python.[46] Here, we considered a specific example of shift-invariant variational autoencoder model (sh-VAE). As stated, sh-VAE is one of the extensions to vanilla VAE where the shift or translational variability of the data is enforced during model training. The encoding-decoding architecture of sh-VAE is very similar to vanilla VAE, except the latent variables are separated into the conventional latent variables and "special" variables associated with the positional variances of the object of interest in the data. Here, the latent maps consist of latent vectors, sampled from normal distribution, and the translational transformation matrix, where the model learns the transformation matrix in an unsupervised way such that the learned latent maps are invariant to the translational transformation matrix. The loss function of a sh-VAE, $\mathcal{L}_{sh-vae}$, can be mathematically written as Eq. 1

$$\mathcal{L}_{sh-vae} = \varphi + \beta(i)D_{KL}(p(\mathbf{Z}|\mathbf{X})||p(\mathbf{Z})) \tag{1}$$

[a]biswasa@ornl.gov
[b]sergei2@utk.edu

Where $\varphi$ is the reconstruction loss, $D_{KL}(p(.|x)||p(.))$ is the KL divergence, $\beta(i)$ is the continuous scale factor of KL divergence at $i^{th}$ training cycle of sh-VAE. The scale factor encourages a better disentanglement of the data, thus provides better learning.[47] Previously, sh-VAE has been implemented for unsupervised discovery of features, patterns, and order parameters from complex experimental microscopic (STM, STEM) images, where key features of several training images have positional variance.[42] However, the sh-VAE model (or any other VAE models) does not consider any physical characteristics of the data, which can reduce the accuracy of learning of such complex images, compromising physics relevant knowledge extraction.

Here, our enhance the performance of the VAE models by introducing inferential bias that allow to encode known physical laws and constraints. Thus, we propose a physics driven VAE (phy-VAE) model to facilitate the training process for the analysis of the atomic and mesoscopic images by combining the existing capability of variational autoencoders to discover the factors of variability in the data under the stated invariances and the new capability of learning any physics-based global criteria to refine the discovery of valuable information from such complex data.

***Proposed Work: Physics-driven VAE -Customizing sh-VAE:***

Figure 1 shows the general approach to integrate the physics driven criteria to a VAE model and develop phy-VAE. As an extension to sh-VAE, the loss function, eq. 2, can be rewritten as below eq. 2.

$$\mathcal{L}_{phy-vae} = \left( \varphi + \beta(i)D_{KL}\left(p(\mathbf{Z}|\mathbf{X})\middle||p(\mathbf{Z})\right) \right) * (w + \Psi) \tag{2}$$

where $\Psi$ is a physic driven loss, $w$ is the slack (a small value) added to new physics driven loss with $0 \leq w \leq 0.5$, to avoid numerical error. In this paper, we derived two loss functions which serves the goal of better learning the phase boundaries with maximizing the smoothness of the learned spatial maps of the encoded latent variables. While the reconstruction and KL divergence loss learn the factors of variability of the data, the new loss component imposes the physical constraints of the data.

[a]biswasa@ornl.gov
[b]sergei2@utk.edu

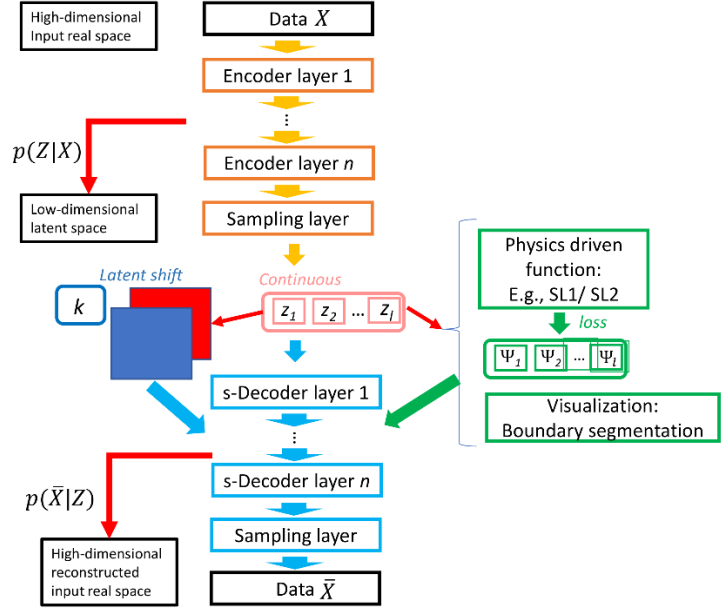

Figure 1. A general approach to Physics-driven VAE.

Here, the first smoothing loss for Ψ is computed through Scharr edge detection filter. As such, it imposes the condition that the total length of the edges within latent image should be minimal, corresponding to the minimization of the interface length due surface tension. The second smoothing loss is computed through peak intensity of the Fourier transform of the spatial latent maps. For the purpose of simplicity, we call these two losses as SL1 and SL2, which we augment to eq 3 as Ψ to minimize. Table 1 and 2 provides the detail algorithm to compute SL1 and SL2.

Here, like the conventional VAE loss computation, these new losses are also calculated at each epoch of training in batches, considering the same batch data ($d \subset D$) in calculating VAE loss. $D$ is the full training data. The batch file, instead of full training data, is considered for computation due to the reasonable memory usage and computational cost of model training for practical implementations. However, due to using the batch file (low resolution latent maps), we approximated SL1 or SL2 instead of exact calculation with full training data (high resolution latent maps). In additional to the loss functions, we have also optionally integrated a visualization function (refer to fig. 1) to track the boundary segmentation of the latent images at the current state of the phy-VAE training. We will discuss more on the boundary segmentation techniques implemented in the result section.

*Table 1: SL1: Edge magnitude of latent images (considering latent dim =2)*

1. Generate the latent maps $Z = (z_1, z_2)$, given the current trained model, $\Lambda$ and data $d$

2. Normalize $Z = (z_1, z_2)$.

3. Use a denoising filter to the latent maps. Here we use bilateral filter with proper tuning as per the data. The reason we choose this filter as it an edge-preserving, denoising filter. Let's annotate the denoised latent maps as $\mathcal{Z} = (z_1, z_2)$

[a]biswasa@ornl.gov
[b]sergei2@utk.edu

4. Find the edge magnitude of $\mathcal{Z} = (z_1, z_2)$. Let's $e_{z_1}, e_{z_2}$ are the edge maps (array) of $(z_1, z_2)$ respectively. Here we used Scharr transform

5. **Calculate loss:** Compute the mean of the sum of the edge maps of denoised latent images as per eq. 3.

$$\Psi = \Psi_{\text{SL1}} = \frac{\sum e_{z_1} + \sum e_{z_2}}{2} \qquad (3)$$

*Table 2: SL2: Peak Intensity of Fourier transform of latent images (considering latent dim =2)*

1. Generate the latent maps $Z = (z_1, z_2)$, given the current trained model, $\Lambda$ and data $d$

2. **Fourier Transform**: Conduct 2D discrete Fourier transform of $(z_1, z_2)$ and shift the zero-frequency component to the center of the spectrum. Let's annotate the Fourier transformed (FFT) latent maps as $\mathcal{F}_Z = (\mathbb{f}_{z_1}, \mathbb{f}_{z_2})$. Do log conversion as $log\mathbb{f}_{z_1} = \log(|\mathbb{f}_{z_1}| + 1)$ and $log\mathbb{f}_{z_2} = \log(|\mathbb{f}_{z_2}| + 1)$

3. Calculate the mean total peak intensity as $pi_{total} = \frac{\sum log\mathbb{f}_{z_1} + \sum log\mathbb{f}_{z_2}}{2}$.

4. Choose the central indices (grid), $[x_{min}, x_{max}], [y_{min}, y_{max}];\ 0 < x_{min}, y_{min} < \frac{i}{2}, i > x_{max}, y_{max} > \frac{i}{2}$, of FFT latent maps and calculate the total central peaks of the FFT latent maps as $\sum_{x_{min}, y_{min}}^{x_{max}, y_{max}} log\mathbb{f}_{z_1}$ and $\sum_{x_{min}, y_{min}}^{x_{max}, y_{max}} log\mathbb{f}_{z_2}$. Here $i$ is the total dimension of square matrix, $d$.

5. Calculate the total intensity outside central peak as $\sum log\mathbb{f}_{z_1} - \sum_{x_{min}, y_{min}}^{x_{max}, y_{max}} log\mathbb{f}_{z_1}$ and $\sum log\mathbb{f}_{z_2} - \sum_{x_{min}, y_{min}}^{x_{max}, y_{max}} log\mathbb{f}_{z_2}$

6. **Calculate loss:** Compute the ratio of the mean outside central peak intensity and mean total peak intensity of FFT latent images as per eq. 4.

$$\Psi = \Psi_{\text{SL2}} = \frac{0.5(\sum log\mathbb{f}_{z_1} - \sum_{x_{min}, y_{min}}^{x_{max}, y_{max}} log\mathbb{f}_{z_1} + \sum log\mathbb{f}_{z_2} - \sum_{x_{min}, y_{min}}^{x_{max}, y_{max}} log\mathbb{f}_{z_2})}{pi_{total}} \qquad (4)$$

In table 1, 2, the process of generating the latent maps (step 1) by VAE is different, depending on the specific VAE model (eg, vanilla VAE, rVAE, sh-VAE etc) we are implementing for a given problem. It is to be noted the phy-VAE is universal to integrate with any VAE models, however, for the demonstration we implemented here with sh-VAE. In this case, for the sh-VAE to encode any positional information of input image patches, we store their original coordinates in the image space and use them to construct a "map" where we assign a value associated with a latent variable z to each coordinate. Here the first two column of the z vector is used to shift the co-ordinate grid by $k\Delta r, \Delta r = z[:, :2], 0 < k \leq 1$, and then the transformed grid is concatenated with the remaining columns of the latent variables before passing to the decoder. The details of the workflow of the latent maps generation in sh-VAE can be found here.[42]

[a]biswasa@ornl.gov
[b]sergei2@utk.edu

## 2. Results

To demonstrate the proposed workflow of phy-VAE, we showcased to different experimental data, such as STEM images of NiO-LSMO heterostructure and BiFeO₃, and STM images of Graphene system. We considered these systems for having two-phase mixture of perovskite LSMO (lanthanum-strontium manganite, LaxSr1-xMnO3) and rock-salt NiO in NiO-LSMO composite, two-phase domain of SrTiO₃ substrate and ferroelectric with ferroic variants, perturbation of electronic structure of graphene on a large number of individual defects. Figure 2 shows the raw experimental images for NiO-LSMO, BiFeO₃ and Graphene. We first train these data with standard sh-VAE models (Appendix fig. A1) to visualize the performance to develop physics relevant latent domain boundaries for reference to compare with the respective latent maps after training with phy-VAEs.

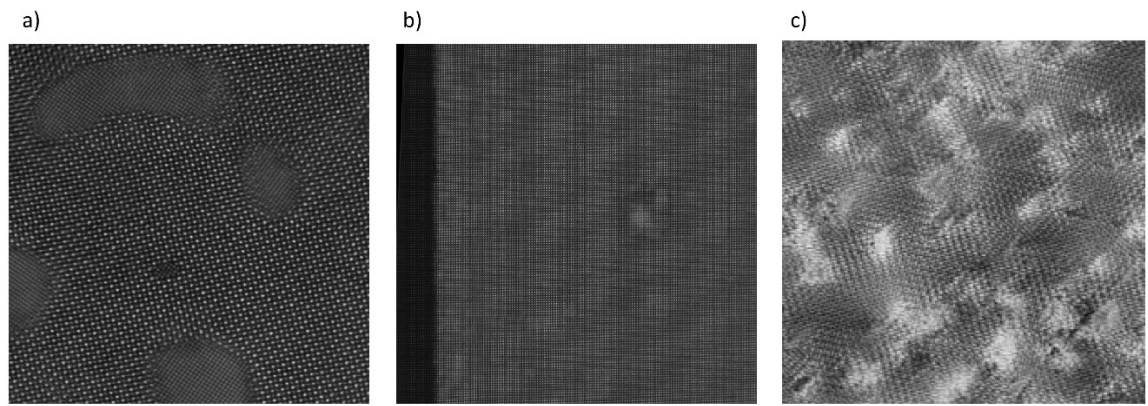

a)     b)     c)

Figure 2. Experimental images of (a) STEM of NiO-LSMO (reprints from fig.6a,[29] -permission information is provided in "Additional Information") , (b) STEM of Sm-doped BiFeO₃ and (c) STM of Graphene Oxide

For NiO-LSMO, we considered denoising factor = 0.1 for SL1, SL2 computation. For model training with either SLI or SL2, we considered 50 epochs for model training with $\beta(i)$ trajectory linearly increasing from 0.05 to 1 as the model completes 50 epochs. In this case, we see minimizing the objectives (SL1 or SL2) on latent map, $z_1$, oversimplifies the learning such that the domain boundary edges disappear. Therefore, we only optimize the latent map, $z_2$, following Table 1, 2 as the phy-VAE model learns $z_1$ efficiently with the existing VAE loss function. After several investigations, we considered the following approach to the visualization (ref Fig 1.) of the image segmentation of the latent maps of the full training data as the training progresses and the final learned latent maps. As we compute the losses with batch data, thus, it is essential to track the performance with whole data. First, we applied a threshold to the latent maps to smooth the full domain if there are negligible difference between the maximum and the minimum latent values. We set this threshold as 1e-3. Once the above threshold is passed, then after normalizing, we denoise the latent maps with bilateral filter, setting the denoising factor = 0.1. Then we applied Otsu threshold and Chan-Vese methods to enhance the boundaries between foreground and background object in the images, and image segmentation respectively. The reason for the selection of these methods, and this sequence of approach to ensure proper defining edges of the latent maps.

[a]biswasa@ornl.gov
[b]sergei2@utk.edu

Figure 3 compares the results of the learned latent maps phy-VAE, at different at different slack values associated to SL1 loss. The same training with phy-VAE models with considering SL2 for physics driven loss is provided in Appendix fig. A2. We can see that sh-VAE yields highly satisfactory results (fig. A1 (a)) in segmenting the phases between NiO and LSMO and therefore the purpose of the phy-VAE is to preserve the existing training of shape reconstruction, without degrading due to enforcing physical constraints. Comparing fig 3, phy-VAE attain the similar phase boundaries of both the latent maps, and while we optimize only the latent map, $z_2$, we see it keeps the actual domain boundary shape and does not affect the boundary shape of $z_1$. Thus, we don't see any sign of distortion or discontinuity over the learned latent manifolds and latent distributions. We find similar performance for the phy-VAE models considering SL2 objective function (fig. A2).

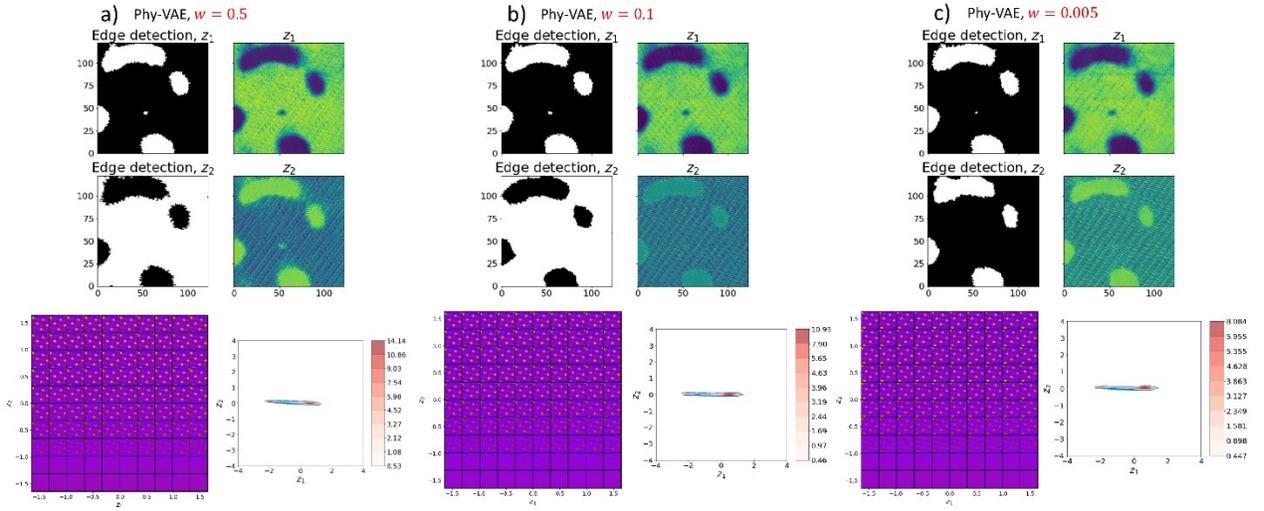

Figure 3. Application to NiO-LSMO system. Comparison between phy-VAEs at different slack values, $w$, of physics driven loss. For each top subfigures, the top and bottom right subplots are learned latent maps $z_1, z_2$ and top and bottom left subplots are edge or phase boundary detection of the respective latent maps through image segmentation. The left and right of the bottom subfigures are the learned latent 2D manifolds and the encoded latent distributions in the latent space respectively for the respective scenarios. In this case, we considered SL1 (Table 1) as the physics driven loss.

Moving forward to more complex BFO system (BiFeO$_3$), during computation of SL1, we considered denoising factor as the standard deviation of the low-resolution latent images. For model training with either SLI or SL2, we considered 50 epochs for model training with $\beta(i)$ trajectory linearly increasing from 0.1 to 1 as the model completes 50 epochs. For the visualization part, we followed the similar approach to the previous case study, except considering denoising factor = 0.3, and considering multi-Otsu threshold instead of binary Otsu and Chan-Vese for the multi-label domain segmentation. From fig. 4, we can now see the limitation of sh-VAE performance as we observe several scattered edges and grains within the phase which deviates the physical behavior that typically order parameters in physical systems are postulated to change smoothly on the length scale of the unit cell. It is to be noted that we followed the same visualization process for image segmentation of the sh-VAE learned latent maps as for the phy-VAE, and thus explain our earlier statement of requirement of enforcing the physics driven knowledge in phy-VAE in spite of the image analysis procedure. Figure 5 shows the latent maps of phy-VAE at different training epochs, considering for SL1, $w = 0.1$. Here the left region (fig. 5f) is the $SrTiO_3$ substrate phase while the large right region is the ferroelectric phase. Within the ferroelectric phase, we see the curved boundary to segment the variants. We can see clearly as the training progress the model reducing and finally eliminating all the small edges and grains within the variants of the ferroelectric segments, following the physical behavior. Figure 6 compares the results of the learned latent maps phy-VAE, at different at different slack values associated to SL1 loss. The same training with phy-VAE models with considering SL2 for physics driven loss is provided in Appendix fig. A3. We clearly see the phy-VAE with both physics driven losses outperformed sh-VAE (fig. 4) in restoring the physical behavior. However, interestingly, we see for fig. 6a, the learned latent map 2 does not detect the ferroic variants while the respective latent map 1 provides well defined boundaries. This can happen due to the fact that latent map 1 learn all the factors of variability of training data in reconstructing the phase, as we can see from the encoded latent distributions of the same. Likewise, for all the cases, we see the learned latent manifolds and distributions does not have any distortion or discontinuities, unlike the case for training with sh-VAE. Finally, we provided figure A4 in Appendix to showcase an example of the regulation of the physics driven loss functions towards the trained latent maps as we maximize those, where we see the increments of boundary edges, distortion, and discontinuities over the latent space.

[a]biswasa@ornl.gov
[b]sergei2@utk.edu

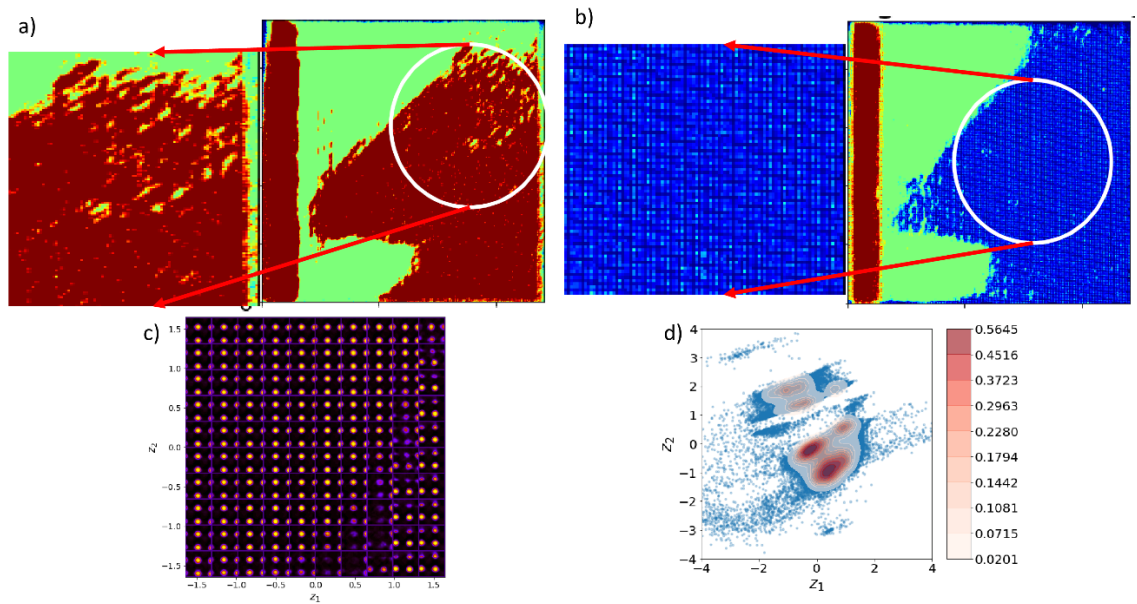

Figure 4. Application of sh-VAE to BFO system. Image segmentation map for edge or phase boundary detection of learned (a) latent map 1, $z_1$ (b) latent map 2, $z_1$. We can see multiple scattered small edges and grains throughout the ferroic variant domain phase. Similarly, we see the (c) distorted learned latent 2D manifolds and (d) discontinuity in the encoded latent distributions in the latent space

[a]biswasa@ornl.gov
[b]sergei2@utk.edu

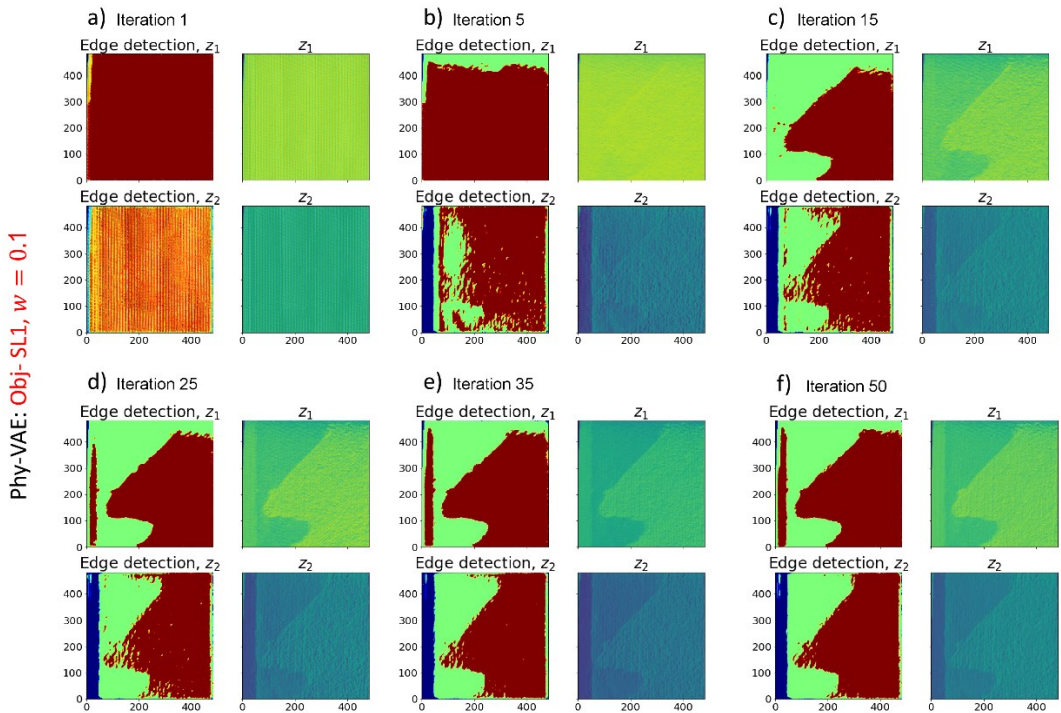

Figure 5. Application to BFO system. Training process of phy-VAE with objective SL1 and $w = 0.1$. For each figure, the top and bottom right subplots are learned latent maps $z_1, z_2$ and top and bottom left subplots are edge or phase boundary detection of the respective latent maps through image segmentation.

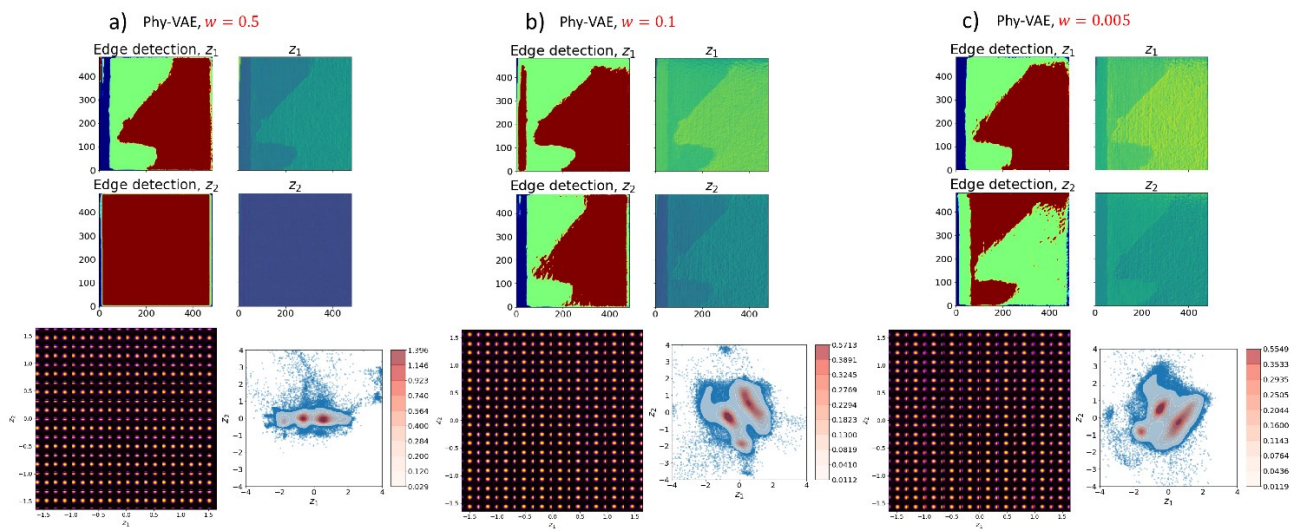

Figure 6. Application to BFO system. Comparison between phy-VAEs at different slack values, $w$, of physics driven loss. For each top subfigures, the top and bottom right subplots are learned latent maps $z_1, z_2$ and top and bottom left subplots are edge or phase boundary detection of the respective latent maps through image segmentation. The left and right of the bottom subfigures are the learned latent 2D manifolds and the encoded latent distributions in the latent space

[a]biswasa@ornl.gov
[b]sergei2@utk.edu

respectively for the respective scenarios. In this case, we considered SL1 (Table 1) as the physics driven loss.

From the application of phy-VAE on previous two material systems, we see the model on one hand perverse the goodness of the phase extraction over the latent maps for both physics driven losses, and on the other hand able to enforce the physical constraints to learn the data in terms of relevant physical behavior of the system. Therefore, we attempt to implement the phy-VAE model into the graphene system where the domain phase is more complex (refer fig 2c). for interpretation. Here, during computation of SL1, we considered the same setup as for BFO system. For model training with either SLI or SL2, we considered 200 epochs for model training with $\beta(i)$ trajectory linearly increasing from 0.01 to 0.5 as the model completes 150 epochs and then kept constant at 0.5 for the remaining 50 training epochs. The image segmentation process also remains the same as for BFO system. Figure 7 shows the results of the learned latent maps with phy-VAE, considering SL1 and SL2 losses. We see the edges and corners of the latent maps has the domain with lowest heatmap, which gradually increases as the distance of latent variables from the edges of latent maps increases, with possibly new domain regions. Similarly, we find no distorted latent manifolds or discontinuous latent distributions.

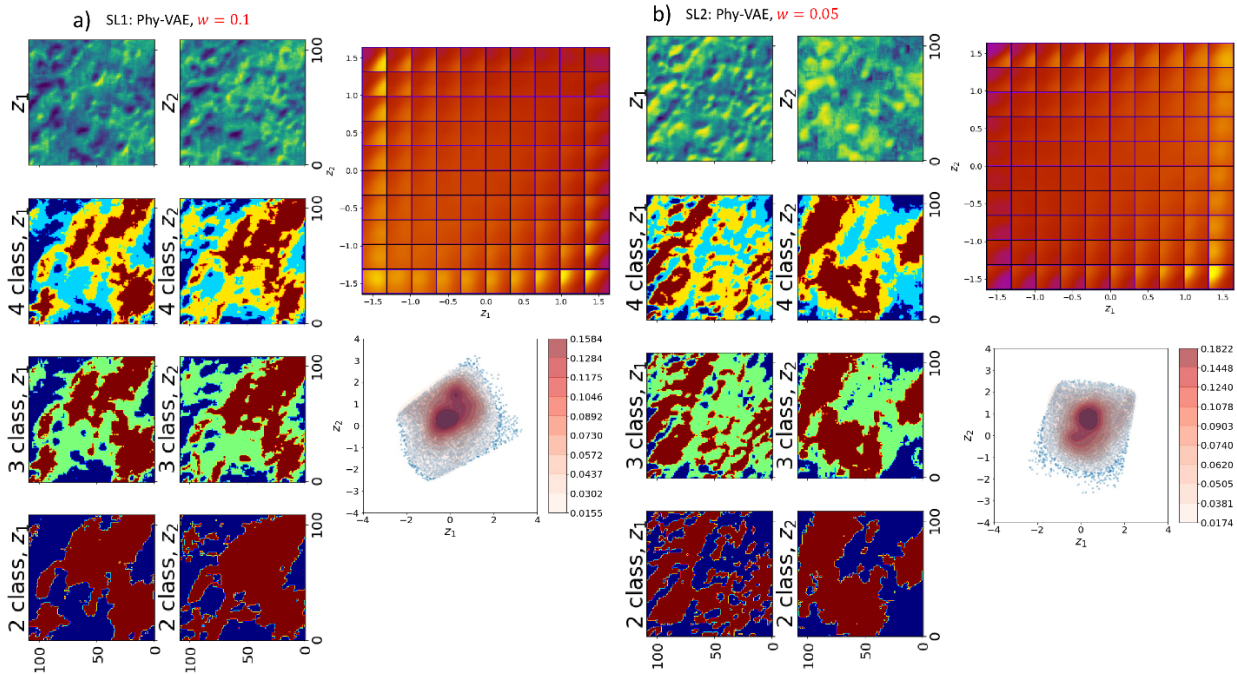

Figure 7. Application to Graphene system. Comparison between phy-VAEs at (a) objective SL1 (Table 1) with $w = 0.1$, (b) objective SL2 (Table 2) with $w = 0.05$ of physics driven loss. For each left subfigures, the left subplots are learned latent map 1, $z_1$, and the respective image segmentation for edge or phase boundary detection with considering 2, 3 and 4 phases (bottom to top). The right subplots are the same for learned latent map 2, $z_2$. The top and bottom of the right subfigures are the learned latent 2D manifolds and the encoded latent distributions in the latent space respectively for the respective scenarios.

[a]biswasa@ornl.gov
[b]sergei2@utk.edu

**Summary**:

We develop a custom VAE, augmenting the physical constraints or knowledge of a material system with the existing VAE training loss, to guide the model training towards an improved (physically meaningful) feature learning of microscopy data analysis. This proposed phy-VAE provides an unsupervised approach for the discovery of order parameter and image segmentation in the presence of other factors of variability. Here this approach is implemented on a shift rotationally invariant variational autoencoder model (sh-VAE) and demonstrated on the open data sets of ferroelectric domains in BiFeO3, LSMO-NiO interface, and graphene. However, this approach to combine the classical VAE KL and reconstruction losses with the physics-based losses representing the known physical inferential biases is universal and can be extended to other custom-design loss functions and other VAE models. This can include more involved sparsity conditions or can also include the non-local conditions and relationships between the latent variables. As such, we believe this approach to be universal and applicable for the unsupervised analysis of the broad variety of physical problems.

**Additional Information:**

See the supplementary material for detailed descriptions and additional figures, related to the research.

Reprints and permission information is available at http://www.nature.com/reprints

**Acknowledgements:**

This work was supported by the US Department of Energy, Office of Science, Office of Basic Energy Sciences, MLExchange Project, award number 107514, and supported by University of Tennessee (Knoxville) start-up funding. The autoencoder research in PyroVED library was supported by the Center for Nanophase Materials Sciences (CNMS), which is a US Department of Energy, Office of Science User Facility at Oak Ridge National Laboratory. Dr. Matthew Chisholm is gratefully acknowledged for the STEM data used in this work

**Conflict of Interest:**

The authors declare no conflict of interest.

**Data Availability Statement:**

The modified code reported here for the purpose of tutorial and application to other data: https://github.com/arpanbiswas52/PaperNotebooks

[a]biswasa@ornl.gov
[b]sergei2@utk.edu

[a]biswasa@ornl.gov
[b]sergei2@utk.edu

[a]biswasa@ornl.gov
[b]sergei2@utk.edu

[a]biswasa@ornl.gov
[b]sergei2@utk.edu

[a]biswasa@ornl.gov
[b]sergei2@utk.edu

Supplementary Materials of the paper titled

**"Combining Variational Autoencoders and Physical Bias for Improved Microscopy Data Analysis"**

Arpan Biswas[1,a], Maxim Ziatdinov[1,2] and Sergei V. Kalinin[3,b]

[1] Center for Nanophase Materials Sciences, Oak Ridge National Laboratory, Oak Ridge, TN 37831
[2] Computational Sciences and Engineering Division, Oak Ridge National Laboratory, Oak Ridge, TN 37831
[3] Materials Science and Engineering, University of Tennessee, Knoxville, TN 37996

## Appendix A. Additional figures

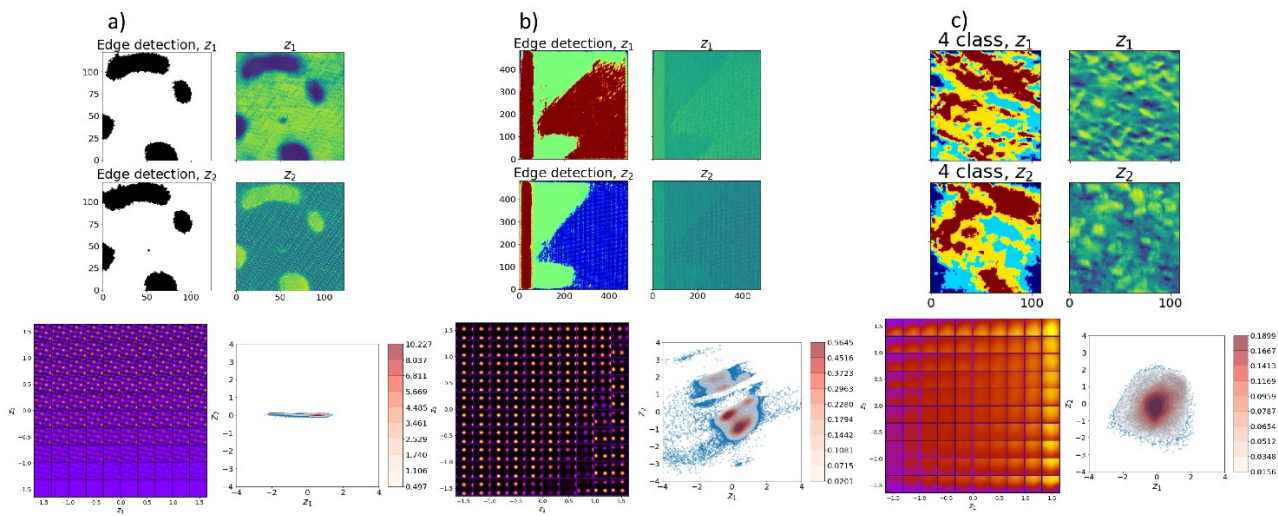

Figure A1. Results from standard sh-VAE model, application to (a) NIO-LSMO, (b) BFO and (c) Graphene system. For each top subfigures, the top and bottom right subplots are learned latent maps $z_1, z_2$ and top and bottom left subplots are edge or phase boundary detection of the respective latent maps through image segmentation. The left and right of the bottom subfigures are the learned latent 2D manifolds and the encoded latent distributions in the latent space respectively for the respective scenarios. As for the graphene data, we derive the latent images into 4 class segmentation since the actual # of classes are unknown.

[a]biswasa@ornl.gov
[b]sergei2@utk.edu

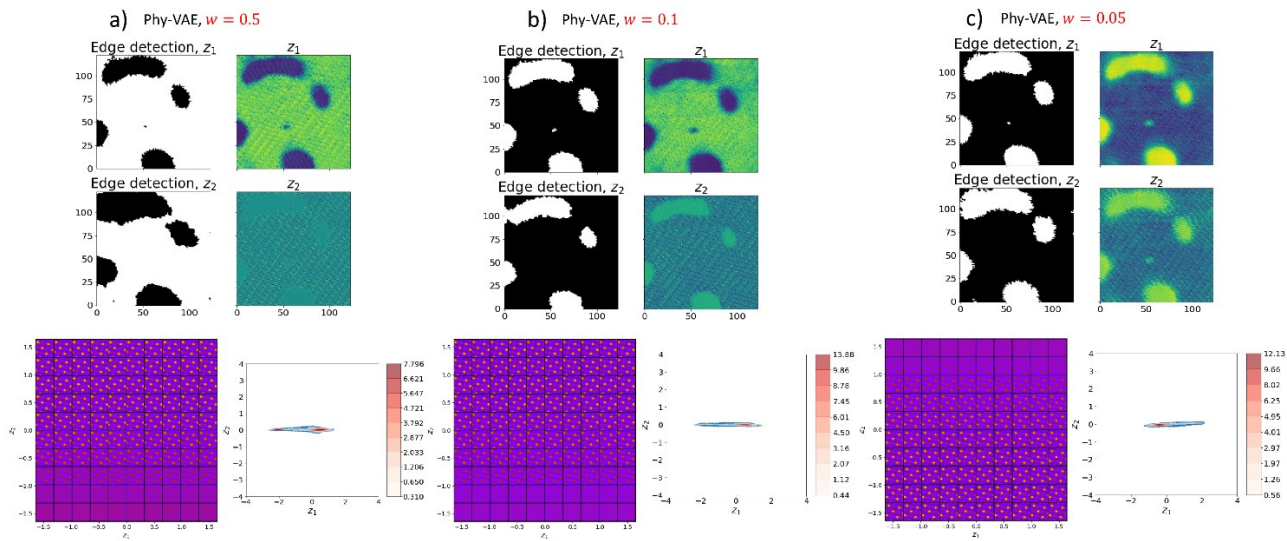

Figure A2. Application to NiO-LSMO system. Comparison between phy-VAEs at different slack values, $w$, of physics driven loss. For each top subfigures, the top and bottom right subplots are learned latent maps $z_1, z_2$ and top and bottom left subplots are edge or phase boundary detection of the respective latent maps through image segmentation. The left and right of the bottom subfigures are the learned latent 2D manifolds and the encoded latent distributions in the latent space respectively for the respective scenarios. In this case, we considered SL2 (Table 2) as the physics driven loss.

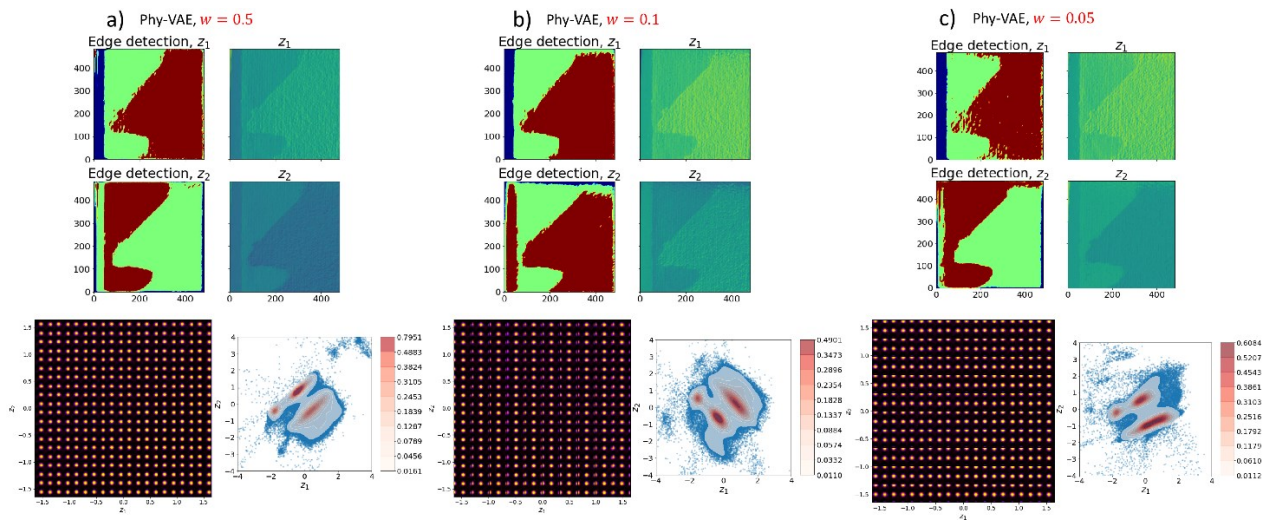

Figure A3. Application to BFO system. Comparison between phy-VAEs at different slack values, $w$, of physics driven loss. For each top subfigures, the top and bottom right subplots are learned latent maps $z_1, z_2$ and top and bottom left subplots are edge or phase boundary detection of the respective latent maps through image segmentation. The left and right of the bottom subfigures are the learned latent 2D manifolds and the encoded latent distributions in the latent space respectively for the respective scenarios. In this case, we considered SL2 (Table 2) as the physics driven loss.

[a]biswasa@ornl.gov
[b]sergei2@utk.edu

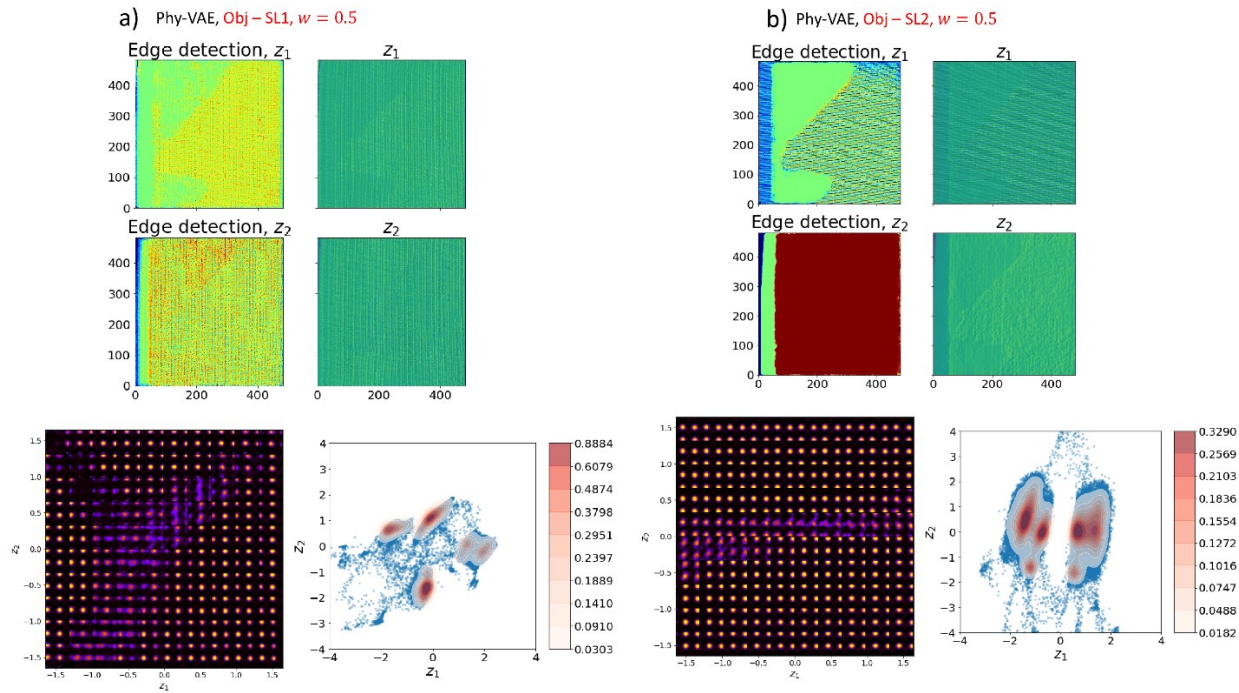

Figure A4. Application to BFO system. Comparison between phy-VAEs at $w = 0.5$, with maximizing the (a) SL1 and (b) SL2 loss functions. Thus, here we want to maximize the domain edges in the latent images. For each top subfigures, the top and bottom right subplots are learned latent maps $z_1, z_2$ and top and bottom left subplots are edge or phase boundary detection of the respective latent maps through image segmentation. The left and right of the bottom subfigures are the learned latent 2D manifolds and the encoded latent distributions in the latent space respectively for the respective scenarios. Similarly to figure 5, we see the increases of edges with the sign of distortion and discontinuity on latent 2D manifolds and encoded latent distributions respectively.

[a]biswasa@ornl.gov
[b]sergei2@utk.edu